\documentclass{article}
\usepackage{iclr2026_conference,times}
\usepackage{amssymb}
\makeatletter
\def\ps@headings{
  \def\@oddhead{}
  \def\@evenhead{}
  \def\@oddfoot{\hfil\thepage\hfil}
  \def\@evenfoot{\hfil\thepage\hfil}
}
\makeatother
\usepackage[colorlinks=true, allcolors=black]{hyperref}
\usepackage{url}
\usepackage{graphicx}
\usepackage{subcaption}
\usepackage{booktabs} 
\usepackage{pifont}
\usepackage{booktabs}   
\usepackage{longtable}
\usepackage{array}
\newcolumntype{L}[1]{>{\raggedright\arraybackslash}p{#1}} 
\usepackage{caption}
\usepackage{multirow}   
\usepackage{siunitx}    

\usepackage{amsmath,amsfonts,bm}

\def\eqref#1{equation~\ref{#1}}

\def\1{\bm{1}}

\DeclareMathAlphabet{\mathsfit}{\encodingdefault}{\sfdefault}{m}{sl}
\SetMathAlphabet{\mathsfit}{bold}{\encodingdefault}{\sfdefault}{bx}{n}

\title{Mechanisms of Non-Monotonic Scaling in\\\mbox{Vision Transformers}}

\iclrfinalcopy

\author{
Anantha Padmanaban Krishna Kumar \\
Department of Computer Science, Boston University\\
\texttt{anantha@bu.edu} \\
}

\begin{document}
\maketitle

\begin{abstract}
Deeper Vision Transformers often perform worse than shallower ones, which challenges common scaling assumptions. Through a systematic empirical analysis of ViT-S, ViT-B, and ViT-L on ImageNet, we identify a consistent three-phase Cliff-Plateau-Climb pattern that governs how representations evolve with depth. We observe that better performance is associated with progressive marginalization of the [CLS] token, originally designed as a global aggregation hub, in favor of distributed consensus among patch tokens. We quantify patterns of information mixing with an Information Scrambling Index, and show that in ViT-L the information-task tradeoff emerges roughly 10 layers later than in ViT-B, and that these additional layers correlate with increased information diffusion rather than improved task performance. Taken together, these results suggest that transformer architectures in this regime may benefit more from carefully calibrated depth that executes clean phase transitions than from simply increasing parameter count. The Information Scrambling Index provides a useful diagnostic for existing models and suggests a potential design target for future architectures. All code is available at: \url{https://github.com/AnanthaPadmanaban-KrishnaKumar/Cliff-Plateau-Climb}.
\end{abstract}

\section{Introduction}
\label{sec:intro}

Vision Transformers have become ubiquitous in computer vision, yet their scaling behavior can defy common intuition. While deeper networks typically yield better representations, we observe the opposite in our setting: ViT-B/16 achieves superior geometric quality compared to the much deeper ViT-L/16. This non-monotonic scaling highlights a gap in our understanding of how transformers process information through depth.

This phenomenon has direct implications for how the field allocates compute. Current practice often assumes that more parameters yield better models, driving massive computational investments. However, if medium-scale models can outperform larger ones in some regimes, this may lead to substantial wasted compute. ViT-L uses substantially more memory, computation, and energy than ViT-B while producing inferior representations in our experiments. Understanding why could help redirect a significant fraction of this compute toward more effective architectures.

\begin{figure}[t!]
    \centering
    \includegraphics[width=0.9\linewidth]{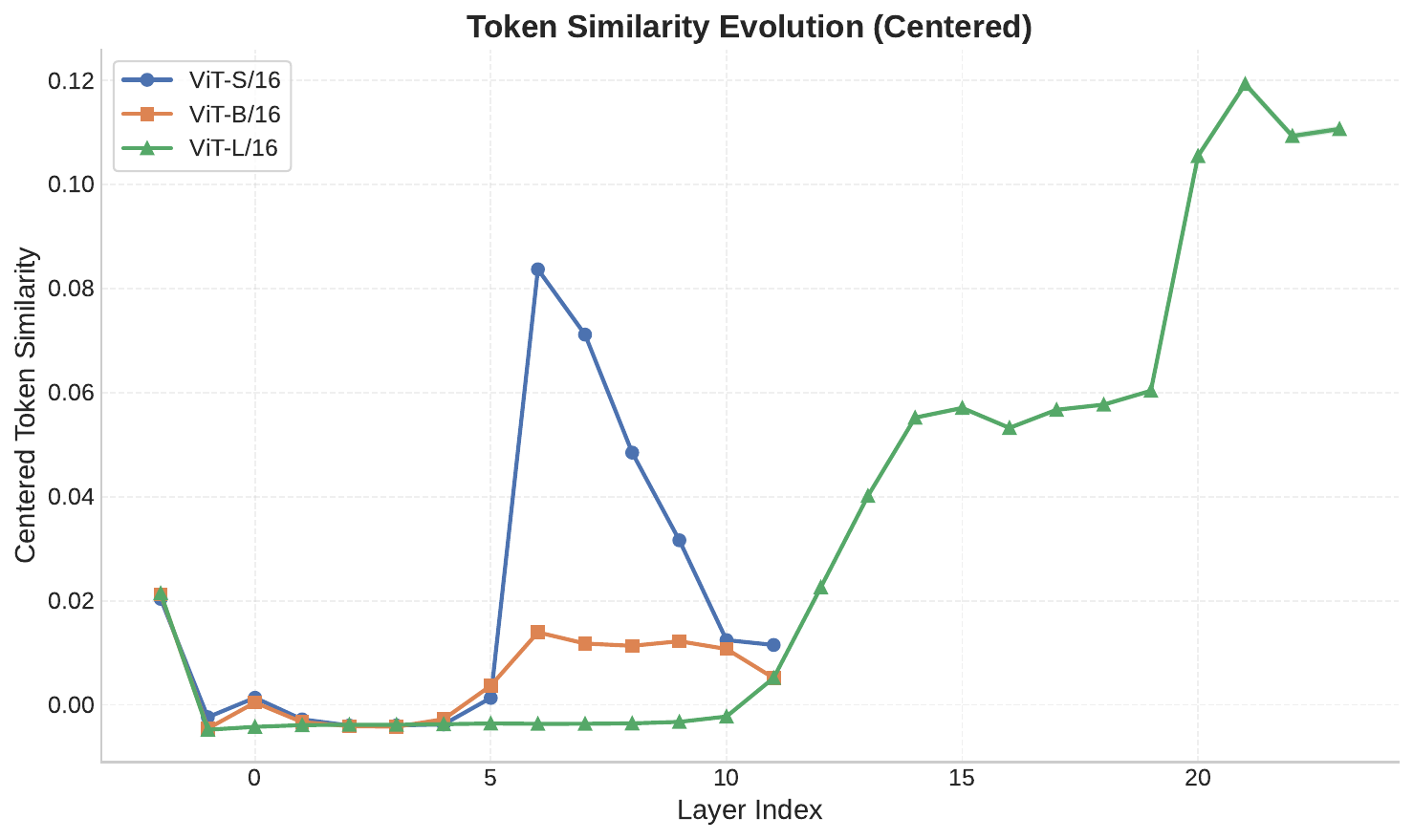}
    \caption{Layer-wise evolution of centered token similarity in Vision Transformers. Three distinct phases emerge consistently across model scales: initial decorrelation (Cliff, layers 0 and 1), extended low-similarity processing (Plateau, middle layers), and terminal re-correlation (Climb, the final few layers, approximately the last three to four). The Plateau duration scales with model depth while maintaining similar similarity ranges.}
    \label{fig:three_phase_pattern}
\end{figure}

To investigate this behavior, we track token representations layer by layer across ViT-S/16, ViT-B/16, and ViT-L/16 on ImageNet. We identify a consistent three-phase Cliff-Plateau-Climb pattern that characterizes how representations evolve with depth (Figure~\ref{fig:three_phase_pattern}). Although all models follow this trajectory, their execution differs: the Plateau extends with depth (6, 12, and 14 layers, respectively), and the \emph{quality} and \emph{timing} of phase transitions are strongly associated with the final representation geometry.

Our most surprising finding concerns the Climb phase, where Neural Collapse emerges in the final layers. The token \texttt{[CLS]}, designed as an information hub, shows systematic marginalization that is correlated with better performance: in ViT-B, the centrality of \texttt{[CLS]} decreases by $\approx 99.2\%$ (from $0.502$ to $0.004$). In our study, networks with the best geometry exhibit patterns consistent with bypassing centralized aggregation in favor of distributed consensus among patch tokens. We quantify patterns of information mixing with an Information Scrambling Index, revealing that ViT-B maintains a controlled mixing regime (roughly $0.004$ to $0.009$) while ViT-L enters a regime of much stronger over-communication (up to $\approx 0.031$), which is associated with degraded geometric quality at greater depth.

Our contributions are:
(1) identification of a consistent Cliff-Plateau-Climb pattern that characterizes Vision Transformer representations in our setting;
(2) empirical evidence that, in this regime, better performance is associated with \texttt{[CLS]} marginalization rather than purely hub-centric aggregation;
(3) introduction of the Information Scrambling Index, which highlights three distinct computational regimes across scale;
(4) empirical observation that, in ViT-L, the tradeoff between information and task performance appears roughly 10 layers deeper than in ViT-B; and
(5) discussion of design principles for depth-efficient transformer architectures motivated by these dynamics.

These findings suggest that, in the Vision Transformers we study, increasing depth primarily extends computation phases rather than introducing qualitatively new features. The Cliff-Plateau-Climb pattern and Information Scrambling Index provide useful diagnostics for analyzing, and potentially designing, efficient architectures that execute clean phase transitions in information flow rather than simply maximizing parameter count.

\section{Related Work}
\label{sec:related_work}

Our work takes a step toward a unified, mechanistic view of non-monotonic scaling in Vision Transformers, connecting three lines of inquiry: the paradox of transformer scaling, layer-wise analysis of internal mechanics, and the geometric theory of Neural Collapse.

\paragraph{The Scaling Paradox and Depth Inefficiency.}
Neural scaling laws suggest that larger models trained on more data tend to achieve better performance \citep{kaplan2020scaling, alabdulmohsin2023getting}, yet the empirical reality for transformers is more complex. Landmark studies have documented performance plateaus and ``double-saturation'' power laws in models with billions of parameters \citep{dehghani2023scaling}. This has led to the observation that compute-optimal model \emph{shape} often matters more than sheer parameter count \citep{alabdulmohsin2023getting}, though the underlying mechanisms remain unclear. Related work has reported symptoms consistent with depth inefficiency, including \emph{attention collapse} \citep{zhou2021deepvit}, \emph{low-pass filtering} \citep{wang2022antioversmoothing}, and \emph{spectral rank collapse} in deep networks \citep{saada2024mind}. A complementary line of work shows that many tokens can be pruned \citep{rao2021dynamicvit} and that final layers often exhibit near-zero activations, suggesting that some of these layers can be removed with limited impact on performance \citep{zhao2024dynamic}. We argue that our Information Scrambling Index provides a common operational lens on these observations, and is consistent with interpreting them as arising from breakdowns in computational strategy rather than from irreducible limitations of depth alone in this regime.

\paragraph{Layer-wise Mechanics and Architectural Assumptions.}
Understanding ViTs has increasingly relied on layer-wise analysis, echoing early findings in language models where information about individual tokens is lost and later recovered across depth \citep{voita2019bottom}. In vision, this perspective has highlighted two key points. First, \emph{positional encodings (PEs) dominate early representations}, causing shallow layers to emphasize spatial location over semantic content \citep{amir2021deep}. Our work interprets the addition of PEs as a functionally critical \emph{decorrelation event} (the ``Cliff'' in Cliff-Plateau-Climb) in our experiments. Second, the conventional role of the \texttt{[CLS]} token as a global information aggregator \citep{dosovitskiy2020image} is increasingly questioned. While pooling strategies have been shown to be effective alternatives \citep{touvron2021training}, we provide empirical evidence, via hub marginalization metrics, that in our setting higher-performing ViTs tend to bypass the \texttt{[CLS]} token as a central bottleneck, shifting toward more distributed aggregation over patch tokens.

\paragraph{The Path to Neural Collapse.}
Neural Collapse (NC) describes a geometric phenomenon often observed in the final-layer features of well-trained classifiers, in which class means converge toward a simplex ETF structure \citep{papyan2020prevalence}. While NC has been extensively studied in CNNs and its theoretical underpinnings continue to evolve \citep{markou2024guiding}, its specific manifestation in Vision Transformers remains comparatively underexplored. Existing work on probing information flow offers tools to observe the \emph{trajectory} toward such collapsed states, but rarely connects the \emph{quality of the computational process} to the \emph{geometry of the endpoint}. Our results help bridge this gap. We find that the Information Scrambling Index, measured in the middle layers, is strongly associated with the final NC geometry in ViT-S/B/L in our experiments, and provides a mechanistic perspective on why the computationally more efficient ViT-B attains a more favorable structure than the deeper, less efficient ViT-L.

\section{Unpacking the Cliff-Plateau-Climb Pattern}
\label{sec:results}

Having introduced the Cliff-Plateau-Climb pattern in Figure~\ref{fig:three_phase_pattern}, we now examine each of its three phases in more detail. We combine layer-wise observations from standard pretrained models with targeted interventions to develop a mechanistic interpretation of each phase. Taken together, these results suggest an end-to-end picture of how Vision Transformers organize representations from initial patch embeddings to final class predictions in our setting.

\subsection{The Cliff: A PE-Driven Decorrelation Event}
\label{sec:cliff}

The first phase of the Cliff-Plateau-Climb pattern is the Cliff: a sharp drop in token similarity is visible in Figure~\ref{fig:three_phase_pattern}. We hypothesize that this decorrelation is primarily driven by positional encoding (PE) addition, with its severity governed by the relative strength of PEs compared to patch embeddings.

Following prior findings that early ViT layers are dominated by positional information \citep{amir2021deep}, we introduce the \textbf{PE Dominance Ratio}, defined as the L2 norm of PE relative to the mean patch-embedding norm, to quantify the strength of this positional signal:
\[
\text{PE-Dom} = \frac{\lVert \text{PE} \rVert_2}{\mathbb{E}\,\lVert z_0 \rVert_2}.
\]

\begin{table}[!htbp]
    \centering
    \caption{PE Dominance increases with model scale and is associated with stronger decorrelation.}
    \label{tab:pe_dominance_effect}
    \begin{tabular}{l ccc | cc}
        \toprule
        & \multicolumn{3}{c}{\textbf{PE strength statistics}} & \multicolumn{2}{c}{\textbf{Centered similarity}} \\
        \cmidrule(r){2-4} \cmidrule(l){5-6}
        \textbf{Model} & \textbf{PE Norm} & \textbf{Patch Norm} & \textbf{Dominance Ratio} & \textbf{Sim($z_0$)} & \textbf{Sim($z_0{+}\text{PE}$)} \\
        \midrule
        ViT-S/16 & 8.107  & 11.697 & 0.693 & 0.020 & -0.002 \\
        ViT-B/16 & 9.530  & 7.918  & 1.204 & 0.021 & -0.005 \\
        ViT-L/16 & 20.090 & 11.864 & 1.693 & 0.021 & -0.005 \\
        \bottomrule
    \end{tabular}
\end{table}

Table~\ref{tab:pe_dominance_effect} shows a scaling relationship: the PE Dominance Ratio climbs from 0.693 (ViT-S) to 1.693 (ViT-L), with stronger PE associated with more severe decorrelation. The decorrelation effect appears to saturate for Base and Large models, suggesting a floor where PE achieves maximal differentiation. This aligns with concurrent work demonstrating that positional-encoding influence varies systematically across model depths \citep{yu2023lape, wu2021ippe}, with recent studies showing that PEs undergo hierarchical transformation from local to global patterns as networks deepen \citep{chu2023conditional}.

\begin{figure}[t!]
  \centering
  \includegraphics[width=0.68\linewidth]{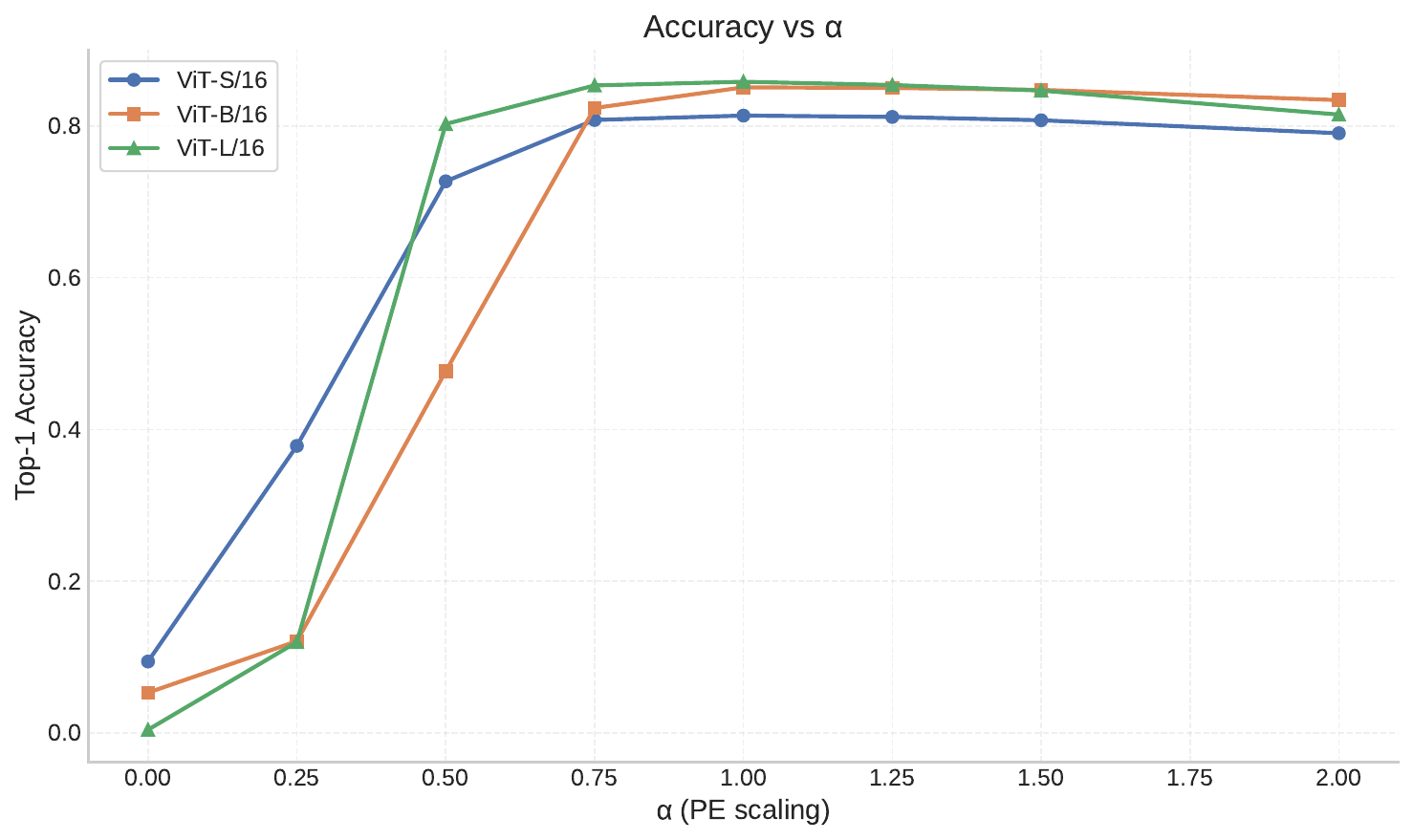}
  \caption{PE strength and model performance. ImageNet top-1 accuracy versus PE scaling factor $\alpha$. All models peak near $\alpha=1.0$, illustrating the functional importance of well-calibrated initial decorrelation.}
  \label{fig:alpha_accuracy}
\end{figure}

To probe this hypothesis more directly, we manually scale PE strength by a factor $\alpha$. Figure~\ref{fig:alpha_accuracy} shows that ImageNet top-1 performance is highly sensitive to this value: accuracy is near chance without PE ($\alpha=0$), peaks near the default setting ($\alpha=1.0$), and degrades when PE is made overly strong, a regime that is approached by ViT-L's native ratio of 1.693. These trends indicate that the Cliff plays an important functional role in these models, rather than being an artifact of initialization.

This parametric scaling provides direct experimental control over the depth of the Cliff and offers evidence consistent with the view that the Cliff is a predictable consequence of standard ViT design and plays a functional role in these models.

\subsection{The Plateau: An Exploration Phase that Scales with Depth}
\label{sec:plateau}

Following the initial Cliff, representations enter the Plateau, which is a long phase of sustained low similarity. We argue that this phase is not stagnation but an important ``working space'' for feature extraction, where in our experiments architectural depth appears to primarily extend this phase's duration. Empirically, we observe that greater depth tends to prolong this low-similarity regime.

\begin{table}[!htbp]
    \centering
    \caption{Plateau length versus model depth. We report the number of consecutive layers where centered similarity remains below a near-zero threshold (see Appendix~\ref{tab:appendix_centered_sim}).}
    \label{tab:plateau_scaling}
    \begin{tabular}{lcc}
        \toprule
        \textbf{Model} & \textbf{Total Layers} & \textbf{Plateau Length (Layers)} \\
        \midrule
        ViT-S/16 & 12 & 6 \\
        ViT-B/16 & 12 & 12 \\
        ViT-L/16 & 24 & 14 \\
        \bottomrule
    \end{tabular}
\end{table}

As shown in Table~\ref{tab:plateau_scaling}, the Plateau length, defined as the number of consecutive layers where centered similarity remains below a near-zero threshold, grows with model size, increasing from 6 layers in ViT-S to 12 in ViT-B and 14 in ViT-L. This suggests that, in our setting, a key role of depth in ViTs is to provide a larger computational budget for feature refinement before final aggregation. The trend is consistent with observations that very deep ViTs have been observed to suffer from attention collapse, where attention maps become progressively similar and eventually identical \citep{zhou2021deepvit}, which has been linked theoretically to the doubly exponential rank collapse of pure attention mechanisms \citep{dong2021attention}. Our measurements provide a quantitative view of this effect: in our setting, greater effective depth is associated with a longer Plateau of diverse representations, which we interpret as supporting more thorough feature refinement before the geometry-forming Climb.

\subsection{The Climb: The Geometric Emergence of the Classifier}
\label{sec:climb}

The final phase of the Cliff-Plateau-Climb pattern, the Climb, is characterized by a rapid re-correlation of tokens in the network's terminal layers. We interpret this not as an arbitrary artifact but as being consistent with the layer-by-layer emergence of the collapsed classifier structure described by Neural Collapse (NC) theory \citep{papyan2020prevalence}. This theory posits that feature representations of samples within the same class collapse to their class mean, with these means arranging into a maximally separable simplex Equiangular Tight Frame (ETF).

\begin{figure}[t!]
    \centering
    \includegraphics[width=\linewidth]{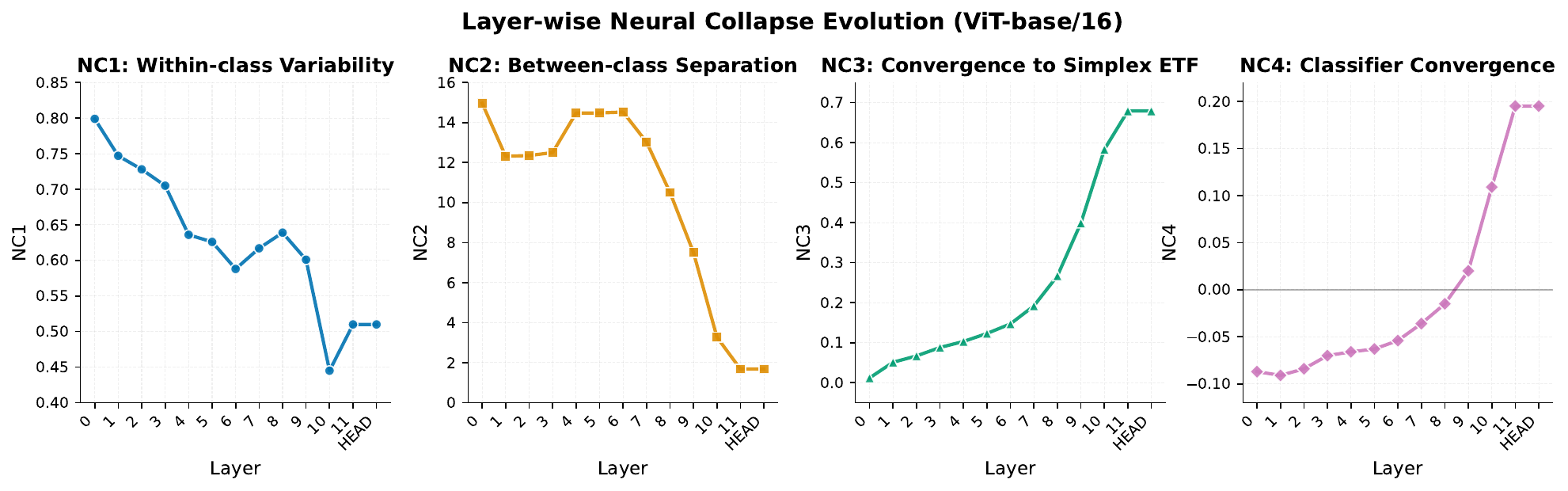}
    \caption{Emergence of Neural Collapse in ViT-Base. Across the final layers, we observe sharp improvement in geometric optimality. NC1 (within-class variance) and NC2 (ETF gap) drop, while NC3 (classifier alignment) and NC4 (decision margin) rise, in a manner consistent with a rapid transition toward a collapsed state.}
    \label{fig:nc_emergence}
\end{figure}

To illustrate this process, we track canonical NC metrics across layers of a representative ViT-Base model. As shown in Figure~\ref{fig:nc_emergence}, the final layers exhibit a pronounced change: metrics measuring geometric optimality (NC1: within-class variance, NC2: ETF gap) improve sharply, while classifier quality metrics (NC3: alignment with class means, NC4: decision margin) increase in tandem. This pattern is consistent with the Climb corresponding to the final consolidation into a collapsed geometric state in this model.

While Neural Collapse has been primarily studied in CNNs \citep{papyan2020prevalence}, related geometric convergence phenomena have been observed in Vision Transformers, such as attention collapse \citep{zhou2021deepvit}, rank collapse in deep layers \citep{dong2021attention,saada2024mind}, and near-zero final layer activation \citep{zhao2024dynamic}. Our systematic measurement of NC metrics across ViT scales reveals a surprising result: geometric optimality does not improve monotonically with model size. As shown in Table~\ref{tab:non_monotonic_scaling}, mid-sized ViT-Base achieves a lower NC2 (closer to the ETF ideal) than both ViT-Small and the much larger ViT-Large.

\begin{table}[!htbp]
    \centering
    \caption{Non-monotonic scaling of classifier geometry. We report Neural Collapse metrics for final layer representations. Lower is better for NC1/NC2; higher for NC3/NC4. ViT-Base achieves the best NC2 score, indicating that its final-layer geometry is closer to the ETF ideal than that of the larger ViT-Large.}
    \label{tab:non_monotonic_scaling}
    \begin{tabular}{lcccc}
        \toprule
        \textbf{Model} & \textbf{NC1} (↓) & \textbf{NC2} (↓) & \textbf{NC3} (↑) & \textbf{NC4} (↑) \\
        \midrule
        ViT-S/16 & 0.546 & 3.420 & 0.775 & 0.117 \\
        ViT-B/16 & 0.510 & \textbf{1.679} & 0.679 & \textbf{0.195} \\
        ViT-L/16 & \textbf{0.468} & 4.173 & \textbf{0.820} & 0.182 \\
        \bottomrule
    \end{tabular}
\end{table}

This analysis supports the view that the Climb corresponds to the formation of a collapsed classifier, but also raises a question about our non-monotonic scaling result: why does mid-sized ViT-Base achieve such favorable geometric structure? Answering this requires moving beyond a purely geometric description to a more mechanistic one. In the following section, we analyze the underlying information flow of the attention graph to better understand the computational processes that appear to underlie this transition, in particular how the trained model effectively bypasses parts of its nominal architectural design.

\section{The Information Plane Reveals Distinct Computational Strategies}
\label{sec:information_plane}

\subsection{The Puzzle: Shared Pattern, Divergent Outcomes}

The Cliff-Plateau-Climb pattern emerges consistently across all three model scales, yet their final geometric quality diverges dramatically: ViT-B achieves an NC2 score of 1.679, substantially better than both ViT-S (3.420) and the much deeper ViT-L (4.173). Thus, while all models share the same coarse representational phase structure, they end in very different geometric regimes. This raises a concrete question: what computational mechanism differentiates how they traverse this shared Cliff-Plateau-Climb landscape?

\subsection{Tracking Information Flow Through Depth}

At each layer $\ell$ we track two complementary, operational quantities: \textbf{task signal} via linear probe accuracy on \texttt{[CLS]}, and \textbf{input-like structure} by reconstructing pre-PE embeddings $z_0$ from patch tokens. These are not mutual informations in the information-theoretic sense, but the performance of specific decoders that serve as probes into what the representation makes linearly accessible.

Two decoder architectures help characterize the network's communication strategy. The \textit{self-only} decoder reconstructs each patch using only its own token:
\[
\hat{z}_0^{(p)} = T_\ell^{(p)} F^\top,
\]
while the \textit{all-to-all} decoder leverages all patches:
\[
\hat{z}_0 = M(T_\ell F^\top)
\quad\text{with } M \in \mathbb{R}^{P \times P},\; F \in \mathbb{R}^{D \times D}.
\]

We normalize reconstruction quality between a perfect oracle (identity) and a null (zeros) decoder:
\begin{equation}
\text{InfoX}_\text{self/all}(\ell)
= 1 - \frac{\text{MSE}_\text{self/all}(\ell) - \text{MSE}_\text{oracle}}{\text{MSE}_\text{null} - \text{MSE}_\text{oracle}}.
\end{equation}
Higher InfoX values indicate that the corresponding decoder family can more faithfully recover $z_0$ from $T_\ell$.

The critical quantity is their difference:
\begin{equation}
\boxed{\text{Information Scrambling Index}(\ell)
= \text{InfoX}_\text{all}(\ell) - \text{InfoX}_\text{self}(\ell)}
\end{equation}

This index measures the added value of cross-patch interactions for reconstructing the initial patch embeddings. Positive values indicate that access to all tokens improves reconstruction over using each token alone; near-zero values indicate that global pooling adds little beyond what is already locally available; negative values indicate over-mixing that is harmful for reconstruction, where exploiting other tokens actually hurts reconstruction compared to the self-only decoder.

\begin{figure}[t!]
    \centering
    \includegraphics[width=\linewidth]{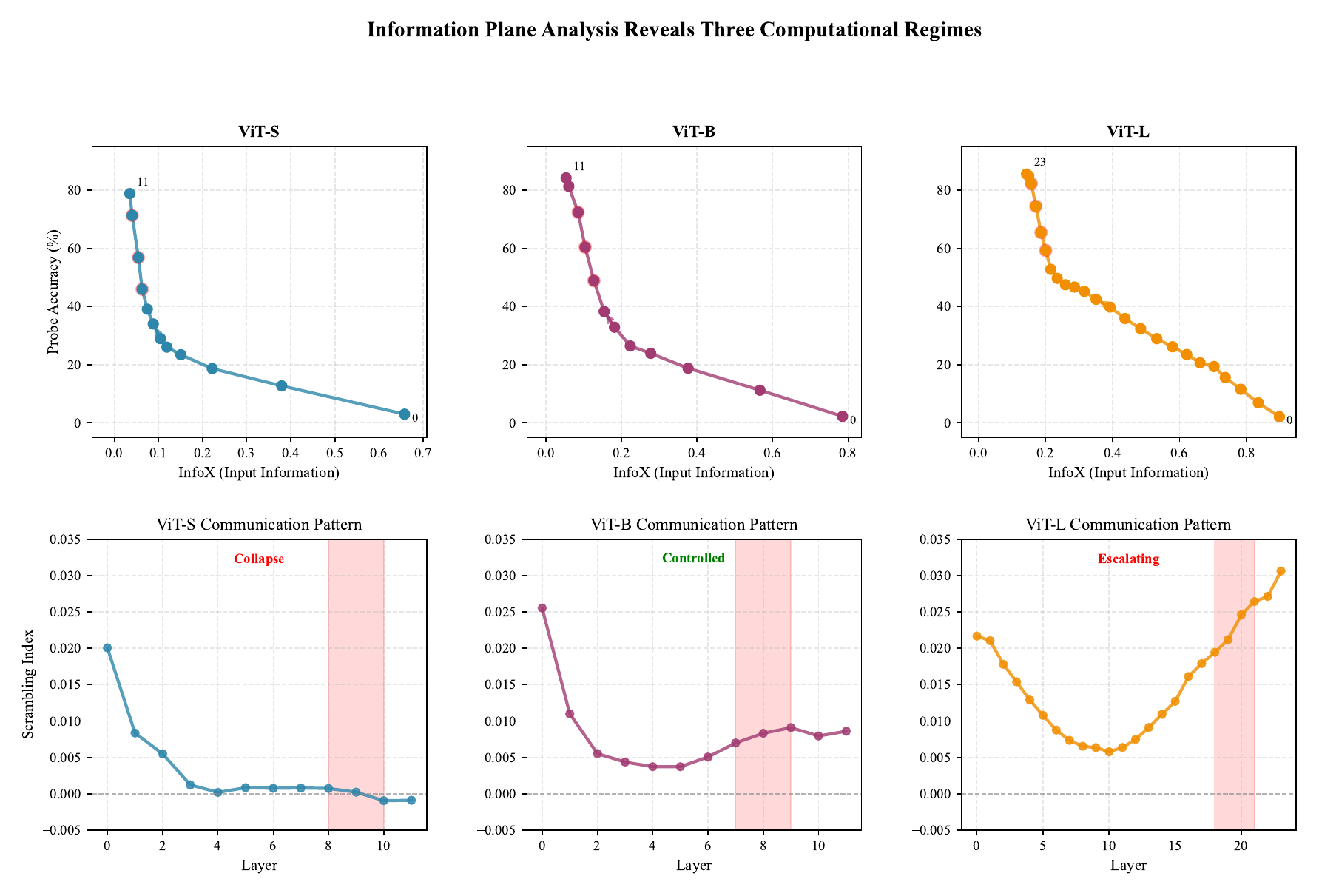}
    \caption{\textbf{Information Plane Analysis.} Panels (a), (b), and (c): all models trade off pre-PE patch structure (InfoX) against task signal but at different rates. ViT-B shows a sharp pivot around layer 8, whereas ViT-L changes more gradually up to about layer 18. Panels (d), (e), and (f): the Scrambling Index reveals their communication regimes. ViT-S exhibits communication collapse, ViT-B maintains controlled mixing, and ViT-L escalates into over-scrambling. Shaded regions mark pivot zones.}
    \label{fig:ip_fingerprints}
\end{figure}

\subsection{Three Distinct Computational Regimes}

Our analysis identifies three regimes with characteristic scrambling dynamics (Figure~\ref{fig:ip_fingerprints}, Table~\ref{tab:three_regimes}). We report the Scrambling Index range over the transformer blocks and summarize the trajectory, task probe gain, and resulting geometry.

\begin{table}[!htbp]
\centering
\small
\caption{Computational regimes characterized by the Information Scrambling Index. Task Gain is the increase in \texttt{[CLS]} probe accuracy from the first to the last layer.}
\begin{tabular}{lccccc}
\toprule
\textbf{Regime} & \textbf{Model} & \textbf{Scrambling Range} & \textbf{Trajectory} & \textbf{Task Gain} & \textbf{NC2} \\
\midrule
Local Processing      & ViT-S & -0.001 to 0.008 & Collapsing   & 32.2\% & 3.420 \\
Controlled Consensus  & ViT-B & 0.004 to 0.009  & Stable       & 34.1\% & \textbf{1.679} \\
Chaotic Diffusion     & ViT-L & 0.007 to 0.031  & Escalating   & 29.5\% & 4.173 \\
\bottomrule
\end{tabular}
\label{tab:three_regimes}
\end{table}

\textbf{ViT-S} exhibits \emph{communication collapse}. After an initially moderate scrambling value (around $0.020$ in the earliest layers), the index drops by roughly 60\% and then remains in a narrow band near zero, eventually turning slightly negative. In this regime, global mixing provides little or even harmful benefit for reconstructing $z_0$, leading to predominantly local per-patch decisions. The model still achieves a nontrivial probe gain (+32.2\%), but does not appear to coordinate sufficiently to reach a strong Neural Collapse geometry.

\textbf{ViT-B} maintains \emph{controlled consensus}. Its Scrambling Index stays in a low but consistently positive band (approximately $0.004$ to $0.009$), with a notable uptick during layers 7 to 9. In this range, global interactions are consistently observed to improve reconstruction over self-only decoding without drifting into destructive or chaotic mixing. ViT-B achieves the highest pivot efficiency: a 34.1\% task probe gain while InfoX drops from 0.127 to 0.085, and simultaneously converges to the best geometric quality (lowest NC2).

\textbf{ViT-L} enters a regime of \emph{chaotic diffusion}. Its Scrambling Index increases monotonically through depth, rising about $4.4\times$ from layer 12 (0.007) to layer 23 (0.031). In this high-scrambling regime, cross-patch mixing is very strong, but the additional layers yield only marginal probe improvements (+1.3\%) while degrading the final Neural Collapse geometry. In our interpretation, the network expends depth maintaining and re-organizing an over-scrambled representation rather than making substantial progress along the tradeoff between information retention and task performance.

\subsection{The Cost of Computational Inefficiency}

In our measurements, ViT-L reaches a comparable information-task tradeoff only after 18 layers, whereas ViT-B does so by layer 8. At layer 8, ViT-L retains InfoX $= 0.532$ with only 28.9\% probe accuracy, while ViT-B has already reached InfoX $= 0.104$ with 60.4\% accuracy. In other words, ViT-L uses roughly 10 additional layers (about 125\% more depth) to reach a comparable checkpoint on the InfoX/Task plane. Throughout this interval, the Scrambling Index steadily escalates, which is consistent with the view that these extra layers are spent predominantly on re-coordinating an increasingly mixed representation rather than driving the model to a qualitatively better tradeoff or geometry.

\subsection{Implications for Transformer Design}

The Information Scrambling Index can serve as a diagnostic for how effectively depth is used to coordinate token interactions. In our ImageNet experiments, more efficient models (such as ViT-B) maintain a low but consistently positive scrambling band (roughly 0.004 to 0.009), where global mixing helps but does not overwhelm local structure. When the index drifts toward zero or negative values, as in ViT-S, global communication either fails to help or becomes mildly destructive, yielding poor geometry despite reasonable accuracy. When it escalates to much larger values through depth, as in ViT-L (up to approximately 0.03), the model enters an over-scrambled regime where additional depth yields diminishing returns.

These numerical thresholds are specific to our normalization and setup, but the qualitative pattern is robust in our experiments: collapse of scrambling is associated with local, uncoordinated computation; stable low-positive scrambling is associated with efficient consensus; escalating high scrambling is associated with unproductive diffusion. The stark inefficiency of ViT-L, which in our experiments needs around ten additional layers to reach the same operational checkpoint as ViT-B, suggests that depth without controlled coordination can end up reorganizing information rather than improving the underlying computation. As such, the Scrambling Index offers a practical lens for diagnosing, in our setting, how transformers allocate depth between useful computation and redundant coordination.

\section{The Mechanism of the Climb: Distributed Consensus via Hub Marginalization}
\label{sec:mechanism}

In our ImageNet Vision Transformers, high performance coincides with the trained model appearing to effectively bypass its own \texttt{[CLS]} token, the architectural hub designed for information aggregation. Through graph-theoretic analysis of attention dynamics across 50{,}000 ImageNet validation images, we find that, in our experiments, the network progressively marginalizes this hub in favor of distributed consensus among patch tokens.

\subsection{Quantifying Attention Dynamics}

We model each layer's attention operation as a directed graph where tokens are nodes and attention weights define edges. From attention weights $\mathbf{A} \in \mathbb{R}^{B \times H \times N \times N}$, we compute: (1) the row-stochastic transition matrix $\mathbf{P}$ by averaging across heads and normalizing, (2) its stationary distribution $\pi$ via power iteration on $\mathbf{P}^\top$, (3) the \textbf{Attention Consensus Index}
\[
\text{ACI}
= 1 - \sigma_2\big(\mathbf{\Pi}^{1/2}\mathbf{P}\mathbf{\Pi}^{-1/2}\big),
\]
where $\mathbf{\Pi} = \mathrm{diag}(\pi)$ and $\sigma_2(\cdot)$ is the second-largest singular value of the $\mathbf{\Pi}$-symmetrized chain, measuring global mixing rate, and (4) the \textbf{CLS Centrality} $\text{CCC} = \pi_0$ quantifying hub importance. Higher ACI indicates faster information propagation; lower CCC indicates hub marginalization.

\subsection{Progressive Marginalization Across Scales}

All three models exhibit progressive \texttt{[CLS]} marginalization but with different outcomes. ViT-S reduces CCC from 0.458 to 0.018 (96.1\%), ViT-B from 0.502 to 0.004 (99.3\%), and ViT-L from 0.373 to 0.010 (97.2\%). The critical distinction lies not in reduction magnitude but in stability and timing. ViT-B achieves a low-hub regime (CCC $< 0.005$) by layer 9, while ViT-S never reaches this threshold. Most strikingly, ViT-L exhibits an instability at layer 20 where CCC spikes to 0.049, an order-of-magnitude increase relative to its neighboring layers, before recovering. This spike coincides with the model that ultimately attains the worst NC2 geometry among the three.

\subsection{The Information Pivot in ViT-B}

\begin{figure}[t!]
    \centering
    \includegraphics[width=0.85\linewidth]{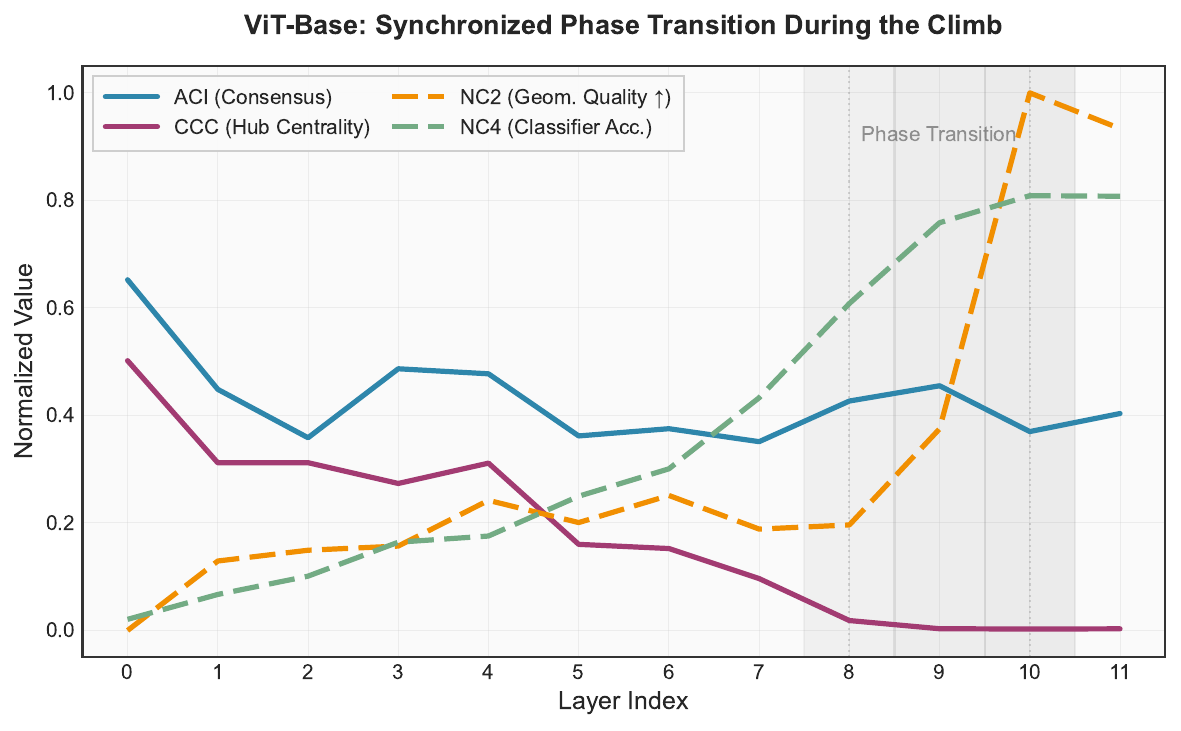}
    \caption{\textbf{Coordinated reorganization in ViT-Base.} During layers 8 to 10 (shaded), hub marginalization (CCC $\downarrow$) and consensus building (ACI $\uparrow$) coincide with improved geometric quality (NC2 $\downarrow$, inverted) and accuracy (NC4 $\uparrow$). Metrics normalized to the range $[0, 1]$.}
    \label{fig:phase_transition}
\end{figure}

Layers 8 to 10 in ViT-B mark a pronounced transition that is consistent with the emergence of distributed consensus. During this ``information pivot'' (Figure~\ref{fig:phase_transition}), CCC drops by roughly 40\% (from $\approx 0.007$ to $\approx 0.004$) while ACI increases by 35\% (0.671 to 0.905), coinciding with NC2 improving from 0.732 to 0.464 and NC4 rising from 0.608 to 0.809. This coordinated reorganization supports the view that Neural Collapse in ViT-B emerges through distributed consensus rather than purely centralized aggregation: the trained model effectively bypasses its architectural bottleneck and relies more heavily on patch-to-patch interactions.

\subsection{Non-Monotonic Scaling and Computational Inefficiency}

Table~\ref{tab:hub_scaling} helps explain why deeper models perform worse in our setting. While all models achieve high consensus (ACI $> 0.91$), marginalization quality differs substantially. The correlation between CCC and NC2 (Spearman $\rho \approx 0.89$) suggests that marginalization quality, rather than consensus speed alone, is more predictive of geometric organization. ViT-L's superior accuracy (NC4 = 0.876) despite inferior geometry (NC2 = 0.541) is consistent with ViT-L relying more on parameter capacity than on efficient representation.

\begin{table}[!htbp]
    \centering
    \caption{Final-layer metrics. ViT-B achieves the best geometry through more complete hub marginalization.}
    \label{tab:hub_scaling}
    \small
    \begin{tabular}{lcccc}
        \toprule
        \textbf{Model} & \textbf{ACI} $\uparrow$ & \textbf{CCC} $\downarrow$ & \textbf{NC2} $\downarrow$ & \textbf{NC4} $\uparrow$ \\
        \midrule
        ViT-S/16 & 0.937 & 0.018 & 0.564 & 0.821 \\
        ViT-B/16 & 0.914 & \textbf{0.004} & \textbf{0.486} & 0.808 \\
        ViT-L/16 & \textbf{0.955} & 0.010 & 0.541 & \textbf{0.876} \\
        \bottomrule
    \end{tabular}
\end{table}

The hub marginalization pattern is closely aligned with the Information Scrambling Index (Section~\ref{sec:information_plane}): ViT-S's collapsed scrambling (down to approximately $-0.001$) is associated with incomplete marginalization (CCC = 0.018), ViT-B's controlled scrambling (roughly 0.004 to 0.009) is associated with more complete marginalization (CCC = 0.004), while ViT-L's excessive scrambling (up to approximately 0.031) is accompanied by unstable marginalization. This relationship suggests that scrambling and marginalization are complementary views of the same underlying reorganization, redistributing information away from the architectural hub into a distributed patch-token consensus.

\subsection{Implications for Transformer Design}

Our findings highlight an inefficiency in this regime: the \texttt{[CLS]} token tends to become a bottleneck that trained models effectively bypass. In our ImageNet ViTs, ViT-B enters a low-hub regime (CCC $\lesssim 0.005$) by about layer 9, whereas ViT-L does not reach comparable marginalization until roughly layer 17, using almost twice as many blocks for a worse final geometry. This mirrors the approximately 125\% depth overhead observed in the Information Plane (Section~\ref{sec:information_plane}), where ViT-L also needs about 18 layers to reach an InfoX/Task tradeoff that ViT-B achieves in 8.

These results suggest that future architectures should provide explicit support for distributed decision-making rather than forcing models to overcome a centralized design. Monitoring CCC during training could help identify inefficient depth utilization early: models that fail to marginalize the hub (high CCC) or that exhibit late-stage spikes in CCC, as in ViT-L, appear to squander depth on re-organizing information rather than improving geometry. The success of mid-sized models is thus consistent with the idea that they possess just enough depth to complete hub marginalization and reach a stable consensus regime without the instability that can arise in larger models.

\section{Discussion}
\label{sec:discussion}

Our work takes a step toward a unified characterization of information processing in Vision Transformers through three complementary lenses: the Cliff-Plateau-Climb pattern revealing geometric evolution, the Information Plane with the Scrambling Index quantifying computational strategies, and graph-theoretic analysis capturing hub marginalization. Together, these lenses give an operational picture of how ViTs transform pixels into predictions in our setting.

Our central observation is a scaling pattern that helps explain why larger models are not automatically better. ViT-B achieves superior geometric quality with what appears to be greater computational efficiency: its controlled Scrambling Index (roughly 0.004 to 0.009) is consistent with a stable form of democratization, where there is sufficient peer-to-peer communication for consensus without entering a strongly mixed regime. Smaller models tend to exhibit communication collapse, leading to predominantly local computation. Larger models tend to exhibit over-communication, spending depth on unproductive re-mixing. Across these regimes, the Scrambling Index tracks behavior and performance: values that are too low indicate under-democratization, while sustained high values correspond to over-diffusion and unstable hub marginalization.

These patterns challenge simple scaling assumptions. In our measurements, ViT-L reaches a comparable information state only after 18 layers, whereas ViT-B does so by layer 8; those extra 10 layers appear to add additional mixing with limited gains in geometry. This suggests that architectures should be optimized for computational strategy, not just parameter count, and that the Scrambling Index and hub metrics could, in principle, act as regularization or early-warning signals during training.

\subsection{Limitations}

Our analysis focuses on pretrained ViT variants (ViT-S/B/L) on ImageNet-1k classification. Extending these patterns to other architectures, tasks, modalities, and training regimes remains future work. The ``optimal'' scrambling range we report is empirically observed in this setting and depends on our proxies (InfoX, ACI, CCC); a full theoretical justification is still open. Moreover, we analyze frozen models and do not intervene in training, so our claims are correlational rather than causal. Nonetheless, we believe that the three-lens framework of geometry, information dynamics, and attention topology provides concrete diagnostics and design signals for transformer architectures.

\section{Conclusion}
\label{sec:conclusion}

We presented an empirical characterization of how Vision Transformers transform pixels into predictions. Across ViT-S/B/L, we find that, in our experiments, these transformers achieve better performance not primarily through centralized aggregation via \texttt{[CLS]}, as the architecture might suggest, but through an effective form of distributed consensus among patch tokens. The Information Scrambling Index indicates that ViT-B outperforms both smaller and larger variants through controlled peer-to-peer communication, while ViT-S shows weak global communication and ViT-L exhibits strong over-communication. In these models, the dominant difference appears to be less about raw capacity and more about computational strategy.

These observations challenge the view that scaling primarily means adding parameters or depth. Our results instead suggest architectures with carefully calibrated depth that execute clean phase transitions in information flow. As the field pushes toward ever-larger models, our findings offer both a cautionary note and a path forward: larger models are not necessarily better when they disrupt the choreography of distributed information flow, and progress may hinge less on sheer scale and more on how transformers coordinate, scramble, and ultimately collapse information into a decision.

\bibliographystyle{iclr2026_conference}
\bibliography{references}

\clearpage
\appendix
\onecolumn

\section{Raw vs. Centered Layer-wise Token Similarity}
\label{app:similarity_tables}

\small
\setlength{\tabcolsep}{4pt}
\renewcommand{\arraystretch}{0.95}

This section provides the full numerical data for the layer-wise evolution of token similarity, supporting the Cliff-Plateau-Climb pattern introduced in Figure~\ref{fig:three_phase_pattern}. We present two metrics: raw token similarity (Table~\ref{tab:appendix_raw_sim}) and centered token similarity (Table~\ref{tab:appendix_centered_sim}).

Raw token similarity measures the average pairwise cosine similarity directly. However, in deep models like ViTs, token representations are often \textbf{anisotropic}, meaning they occupy a narrow cone in the embedding space. This results in a high baseline similarity for all tokens, which can obscure more subtle layer-wise dynamics.

To correct for this, we compute \textbf{centered token similarity}. In this metric, the mean token representation for a given image (i.e., the ``center of the cone'') is subtracted from each patch token before calculating the average pairwise cosine similarity. As the comparison between Table~\ref{tab:appendix_raw_sim} and Table~\ref{tab:appendix_centered_sim} shows, this simple correction removes the high, confounding baseline and more clearly reveals the Cliff-Plateau-Climb pattern. The data from Table~\ref{tab:appendix_centered_sim} is plotted in Figure~\ref{fig:three_phase_pattern} of the main paper.

\begin{longtable}{@{}rlccc@{}}
\caption{Raw Token Similarity. Average pairwise cosine similarity of patch tokens at each layer for ViT-S/16, ViT-B/16, and ViT-L/16. Values are reported as mean with a 95\% bootstrapped confidence interval across the ImageNet validation set. Note the consistently high similarity values across most layers, which is characteristic of representational anisotropy.}
\label{tab:appendix_raw_sim}
\\
\toprule
\textbf{Layer Index} & \textbf{Layer Name} & \textbf{ViT-S/16 (Mean [CI])} & \textbf{ViT-B/16 (Mean [CI])} & \textbf{ViT-L/16 (Mean [CI])} \\
\midrule
\endfirsthead
\toprule
\textbf{Layer Index} & \textbf{Layer Name} & \textbf{ViT-S/16 (Mean [CI])} & \textbf{ViT-B/16 (Mean [CI])} & \textbf{ViT-L/16 (Mean [CI])} \\
\midrule
\endhead
\midrule \multicolumn{5}{r}{\textit{Continued on next page}}\\
\endfoot
\bottomrule
\endlastfoot
-2 & z0             & 0.325 [0.324, 0.326] & 0.540 [0.539, 0.541] & 0.295 [0.294, 0.296] \\
-1 & z0\_plus\_pe   & 0.191 [0.191, 0.192] & 0.210 [0.210, 0.210] & 0.059 [0.059, 0.059] \\
0  & block\_0       & 0.437 [0.436, 0.438] & 0.489 [0.489, 0.490] & 0.090 [0.090, 0.091] \\
1  & block\_1       & 0.305 [0.305, 0.306] & 0.247 [0.247, 0.247] & 0.101 [0.100, 0.101] \\
2  & block\_2       & 0.227 [0.227, 0.228] & 0.209 [0.209, 0.210] & 0.108 [0.107, 0.108] \\
3  & block\_3       & 0.277 [0.276, 0.277] & 0.224 [0.224, 0.225] & 0.126 [0.125, 0.126] \\
4  & block\_4       & 0.251 [0.251, 0.252] & 0.245 [0.245, 0.246] & 0.166 [0.166, 0.166] \\
5  & block\_5       & 0.260 [0.260, 0.261] & 0.235 [0.235, 0.235] & 0.191 [0.191, 0.191] \\
6  & block\_6       & 0.242 [0.241, 0.242] & 0.238 [0.237, 0.238] & 0.209 [0.209, 0.210] \\
7  & block\_7       & 0.230 [0.230, 0.231] & 0.232 [0.232, 0.232] & 0.208 [0.207, 0.208] \\
8  & block\_8       & 0.247 [0.246, 0.247] & 0.254 [0.253, 0.254] & 0.202 [0.201, 0.202] \\
9  & block\_9       & 0.323 [0.322, 0.323] & 0.308 [0.308, 0.309] & 0.206 [0.206, 0.206] \\
10 & block\_10      & 0.410 [0.409, 0.410] & 0.462 [0.461, 0.463] & 0.203 [0.203, 0.203] \\
11 & block\_11      & 0.517 [0.516, 0.517] & 0.686 [0.685, 0.687] & 0.202 [0.202, 0.203] \\
12 & block\_12      & ---                   & ---                   & 0.198 [0.198, 0.199] \\
13 & block\_13      & ---                   & ---                   & 0.196 [0.195, 0.196] \\
14 & block\_14      & ---                   & ---                   & 0.194 [0.194, 0.195] \\
15 & block\_15      & ---                   & ---                   & 0.200 [0.200, 0.200] \\
16 & block\_16      & ---                   & ---                   & 0.207 [0.207, 0.208] \\
17 & block\_17      & ---                   & ---                   & 0.227 [0.227, 0.228] \\
18 & block\_18      & ---                   & ---                   & 0.252 [0.252, 0.253] \\
19 & block\_19      & ---                   & ---                   & 0.276 [0.276, 0.277] \\
20 & block\_20      & ---                   & ---                   & 0.303 [0.302, 0.303] \\
21 & block\_21      & ---                   & ---                   & 0.345 [0.344, 0.345] \\
22 & block\_22      & ---                   & ---                   & 0.383 [0.382, 0.383] \\
23 & block\_23      & ---                   & ---                   & 0.481 [0.481, 0.482] \\
\end{longtable}

\begin{longtable}{@{}rlccc@{}}
\caption{Centered Token Similarity. Average pairwise cosine similarity of \textbf{mean-centered} patch tokens at each layer. This correction for anisotropy clearly reveals the Cliff--Plateau--Climb dynamic. These values are used to generate Figure~\ref{fig:three_phase_pattern} in the main text.}
\label{tab:appendix_centered_sim}
\\
\toprule
\textbf{Layer Index} & \textbf{Layer Name} & \textbf{ViT-S/16 (Mean [CI])} & \textbf{ViT-B/16 (Mean [CI])} & \textbf{ViT-L/16 (Mean [CI])} \\
\midrule
\endfirsthead
\toprule
\textbf{Layer Index} & \textbf{Layer Name} & \textbf{ViT-S/16 (Mean [CI])} & \textbf{ViT-B/16 (Mean [CI])} & \textbf{ViT-L/16 (Mean [CI])} \\
\midrule
\endhead
\midrule \multicolumn{5}{r}{\textit{Continued on next page}}\\
\endfoot
\bottomrule
\endlastfoot
-2 & z0             & 0.020 [0.020, 0.021] & 0.021 [0.021, 0.022] & 0.021 [0.021, 0.22] \\
-1 & z0\_plus\_pe   & -0.002 [-0.002, -0.002] & -0.005 [-0.005, -0.005] & -0.005 [-0.005, -0.005] \\
0  & block\_0       & 0.001 [0.001, 0.001] & 0.000 [0.000, 0.001] & -0.004 [-0.004, -0.004] \\
1  & block\_1       & -0.003 [-0.003, -0.003] & -0.003 [-0.003, -0.003] & -0.004 [-0.004, -0.004] \\
2  & block\_2       & -0.004 [-0.004, -0.004] & -0.004 [-0.004, -0.004] & -0.004 [-0.004, -0.004] \\
3  & block\_3       & -0.004 [-0.004, -0.004] & -0.004 [-0.004, -0.004] & -0.004 [-0.004, -0.004] \\
4  & block\_4       & -0.004 [-0.004, -0.004] & -0.003 [-0.003, -0.003] & -0.004 [-0.004, -0.004] \\
5  & block\_5       & 0.001 [0.001, 0.001] & 0.004 [0.004, 0.004] & -0.004 [-0.004, -0.004] \\
6  & block\_6       & 0.084 [0.083, 0.084] & 0.014 [0.014, 0.014] & -0.004 [-0.004, -0.004] \\
7  & block\_7       & 0.071 [0.071, 0.071] & 0.012 [0.012, 0.012] & -0.004 [-0.004, -0.004] \\
8  & block\_8       & 0.048 [0.048, 0.049] & 0.011 [0.011, 0.011] & -0.004 [-0.004, -0.004] \\
9  & block\_9       & 0.032 [0.031, 0.032] & 0.012 [0.012, 0.012] & -0.003 [-0.003, -0.003] \\
10 & block\_10      & 0.012 [0.012, 0.013] & 0.011 [0.011, 0.011] & -0.002 [-0.002, -0.002] \\
11 & block\_11      & 0.012 [0.011, 0.012] & 0.005 [0.005, 0.005] & 0.005 [0.005, 0.005] \\
12 & block\_12      & ---                   & ---                   & 0.023 [0.022, 0.023] \\
13 & block\_13      & ---                   & ---                   & 0.040 [0.040, 0.040] \\
14 & block\_14      & ---                   & ---                   & 0.055 [0.055, 0.055] \\
15 & block\_15      & ---                   & ---                   & 0.057 [0.057, 0.057] \\
16 & block\_16      & ---                   & ---                   & 0.053 [0.053, 0.053] \\
17 & block\_17      & ---                   & ---                   & 0.057 [0.056, 0.057] \\
18 & block\_18      & ---                   & ---                   & 0.058 [0.057, 0.058] \\
19 & block\_19      & ---                   & ---                   & 0.060 [0.060, 0.061] \\
20 & block\_20      & ---                   & ---                   & 0.105 [0.105, 0.106] \\
21 & block\_21      & ---                   & ---                   & 0.119 [0.119, 0.120] \\
22 & block\_22      & ---                   & ---                   & 0.109 [0.109, 0.110] \\
23 & block\_23      & ---                   & ---                   & 0.111 [0.110, 0.111] \\
\end{longtable}

\section{Information Plane Analysis Details}
\label{app:ip_methodology}

\subsection{Complete Numerical Results}

Tables~\ref{tab:vits_metrics}--\ref{tab:vitl_metrics_1} present complete layer-wise metrics for all three model scales with 95\% bootstrap confidence intervals over 2000 resamples.

\begin{table}[!htbp]
\centering
\caption{ViT-S/16 Information Plane Metrics. Probe accuracy reported as mean [95\% CI].}
\label{tab:vits_metrics}
\small
\begin{tabular}{rccccccc}
\toprule
Layer & Probe Acc. & InfoX$_{\text{self}}$ & InfoX$_{\text{all}}$ & Scrambling & Task Gain & InfoX Drop \\
\midrule
 0 &  3.00 [2.62, 3.40] & 0.658 & 0.678 & 0.020 & --     & --      \\
 1 & 12.72 [11.94, 13.54] & 0.379 & 0.388 & 0.008 & +9.72  & -0.278  \\
 2 & 18.66 [17.76, 19.54] & 0.222 & 0.227 & 0.005 & +5.94  & -0.157  \\
 3 & 23.42 [22.44, 24.42] & 0.151 & 0.152 & 0.001 & +4.76  & -0.071  \\
 4 & 26.04 [25.00, 27.10] & 0.119 & 0.120 & 0.000 & +2.62  & -0.032  \\
 5 & 28.96 [27.84, 30.04] & 0.105 & 0.106 & 0.001 & +2.92  & -0.015  \\
 6 & 34.00 [32.90, 35.12] & 0.088 & 0.089 & 0.001 & +5.04  & -0.016  \\
 7 & 39.08 [37.92, 40.22] & 0.075 & 0.076 & 0.001 & +5.08  & -0.013  \\
 8 & 45.94 [44.76, 47.12] & 0.063 & 0.064 & 0.001 & +6.86  & -0.012  \\
 9 & 56.82 [55.62, 58.06] & 0.055 & 0.055 & 0.000 & +10.88 & -0.009  \\
10 & 71.30 [70.20, 72.40] & 0.041 & 0.040 & -0.001 & +14.48 & -0.014 \\
11 & 78.82 [77.82, 79.80] & 0.035 & 0.034 & -0.001 & +7.52  & -0.005 \\
\bottomrule
\end{tabular}
\end{table}

\begin{table}[!htbp]
\centering
\caption{ViT-B/16 Information Plane Metrics. Probe accuracy reported as mean [95\% CI].}
\label{tab:vitb_metrics}
\small
\begin{tabular}{rccccccc}
\toprule
Layer & Probe Acc. & InfoX$_{\text{self}}$ & InfoX$_{\text{all}}$ & Scrambling & Task Gain & InfoX Drop \\
\midrule
 0 &  2.26 [1.92, 2.62] & 0.785 & 0.810 & 0.026 & --     & --      \\
 1 & 11.24 [10.50, 11.98] & 0.566 & 0.577 & 0.011 & +8.98  & -0.219  \\
 2 & 18.78 [17.90, 19.72] & 0.376 & 0.382 & 0.006 & +7.54  & -0.190  \\
 3 & 23.90 [22.88, 24.92] & 0.278 & 0.282 & 0.004 & +5.12  & -0.099  \\
 4 & 26.44 [25.38, 27.48] & 0.223 & 0.227 & 0.004 & +2.54  & -0.055  \\
 5 & 32.88 [31.78, 34.00] & 0.182 & 0.185 & 0.004 & +6.44  & -0.042  \\
 6 & 38.28 [37.18, 39.40] & 0.154 & 0.159 & 0.005 & +5.40  & -0.027  \\
 7 & 48.88 [47.66, 50.08] & 0.127 & 0.134 & 0.007 & +10.60 & -0.027  \\
 8 & 60.40 [59.28, 61.56] & 0.104 & 0.112 & 0.008 & +11.52 & -0.023  \\
 9 & 72.38 [71.34, 73.46] & 0.085 & 0.094 & 0.009 & +11.98 & -0.019  \\
10 & 81.28 [80.38, 82.20] & 0.061 & 0.069 & 0.009 & +8.90  & -0.025  \\
11 & 84.20 [83.32, 85.00] & 0.054 & 0.062 & 0.009 & +2.92  & -0.007  \\
\bottomrule
\end{tabular}
\end{table}

\begin{table}[!htbp]
\centering
\caption{ViT-L/16 Information Plane Metrics. Probe accuracy reported as mean [95\% CI].}
\label{tab:vitl_metrics_1}
\small
\begin{tabular}{rccccccc}
\toprule
Layer & Probe Acc. & InfoX$_{\text{self}}$ & InfoX$_{\text{all}}$ & Scrambling & Task Gain & InfoX Drop \\
\midrule
 0 &  2.12 [1.80, 2.44] & 0.898 & 0.919 & 0.022 & --     & --      \\
 1 &  6.86 [6.26, 7.44] & 0.835 & 0.856 & 0.021 & +4.74  & -0.063  \\
 2 & 11.56 [10.80, 12.30] & 0.782 & 0.800 & 0.018 & +4.70  & -0.052  \\
 3 & 15.60 [14.76, 16.42] & 0.736 & 0.752 & 0.015 & +4.04  & -0.046  \\
 4 & 19.38 [18.42, 20.30] & 0.703 & 0.715 & 0.013 & +3.78  & -0.034  \\
 5 & 20.66 [19.70, 21.62] & 0.661 & 0.672 & 0.011 & +1.28  & -0.042  \\
 6 & 23.50 [22.58, 24.54] & 0.621 & 0.630 & 0.009 & +2.84  & -0.040  \\
 7 & 26.16 [25.12, 27.18] & 0.579 & 0.586 & 0.007 & +2.66  & -0.042  \\
 8 & 28.94 [27.86, 30.02] & 0.532 & 0.538 & 0.007 & +2.78  & -0.047  \\
 9 & 32.38 [31.26, 33.52] & 0.484 & 0.490 & 0.006 & +3.44  & -0.048  \\
10 & 35.86 [34.76, 37.06] & 0.437 & 0.443 & 0.006 & +3.48  & -0.047  \\
11 & 39.80 [38.66, 40.98] & 0.392 & 0.399 & 0.006 & +3.94  & -0.045  \\
12 & 42.48 [41.24, 43.68] & 0.351 & 0.358 & 0.007 & +2.68  & -0.041  \\
13 & 45.24 [44.08, 46.42] & 0.315 & 0.325 & 0.009 & +2.76  & -0.035  \\
14 & 46.68 [45.50, 47.94] & 0.286 & 0.297 & 0.011 & +1.44  & -0.029  \\
15 & 47.48 [46.30, 48.68] & 0.259 & 0.272 & 0.013 & +0.80  & -0.027  \\
16 & 49.68 [48.48, 50.88] & 0.235 & 0.251 & 0.016 & +2.20  & -0.025  \\
17 & 52.76 [51.58, 53.90] & 0.216 & 0.234 & 0.018 & +3.08  & -0.019  \\
18 & 59.28 [58.18, 60.46] & 0.200 & 0.220 & 0.019 & +6.52  & -0.015  \\
19 & 65.46 [64.36, 66.58] & 0.186 & 0.208 & 0.021 & +6.18  & -0.014  \\
20 & 74.48 [73.46, 75.54] & 0.171 & 0.196 & 0.025 & +9.02  & -0.015  \\
21 & 82.24 [81.36, 83.16] & 0.158 & 0.184 & 0.026 & +7.76  & -0.014  \\
22 & 84.88 [84.02, 85.72] & 0.150 & 0.177 & 0.027 & +2.64  & -0.008  \\
23 & 85.52 [84.66, 86.36] & 0.143 & 0.174 & 0.031 & +0.64  & -0.007  \\
\bottomrule
\end{tabular}
\end{table}

\subsection{Methodological Details}

\textbf{Decoder Architectures:}
The self-only decoder uses a linear transformation $F \in \mathbb{R}^{D \times D}$ applied independently to each patch token, where $D = 384$ (ViT-S), 768 (ViT-B), 1024 (ViT-L). The all-to-all decoder combines a linear transformation $F \in \mathbb{R}^{D \times D}$ with a patch mixing matrix $M \in \mathbb{R}^{P \times P}$, where $P = 196$ for all models.

\textbf{Training Configuration:}
Classification probes used a linear layer with bias, Adam optimizer (lr=$10^{-2}$, no weight decay), batch size 8192, early stopping with patience 10, maximum 100 epochs. Reconstruction decoders used Adam optimizer (lr=$3 \times 10^{-3}$, weight decay=$10^{-4}$), batch size 2048, early stopping with patience 10, maximum 200 epochs, MSE loss. All experiments used 80\% train, 10\% validation, 10\% test splits stratified by class.

\textbf{Information Pivot Identification:}
The information pivot is identified as the contiguous set of layers exhibiting maximum rate of task information gain (probe accuracy increase per layer) while simultaneously showing significant input information decrease (InfoX drop greater than 0.01 per layer).

\section{Validation of Probing Methodology}
\label{app:sanity_checks}

To help ensure validity and robustness of our information-related measurements, we conducted control experiments for both classification and reconstruction probes across all three model scales.

\subsection{Classification Probe: Random-Label Control}
\label{app:sanity_cls}

To verify that linear classification probes learn meaningful class-specific signals, we randomly permuted training labels while keeping validation and test sets unchanged. Across all layers and model scales, test accuracies ranged from 0.02\% to 0.20\% (with chance at 0.1\% for 1000 classes), all remaining well below our conservative threshold of 0.3\% ($3\times$ chance). This supports the conclusion that the probes do not exploit spurious correlations or structural artifacts.

\subsection{Reconstruction Decoder: Permutation Control}
\label{app:sanity_recon}

To verify that reconstruction decoders learn spatial correspondence between patch tokens, we trained decoders to reconstruct initial patch embeddings ($z_0$) whose patches were randomly shuffled per image, while input tokens $T_\ell$ remained in their original positions. This spatial disruption severely degraded reconstruction performance, with test MSE approaching the null baseline (mean differences: $-0.027$ for ViT-S, $-0.035$ for ViT-B, $-0.089$ for ViT-L). The slightly-better-than-null performance likely reflects residual statistical structure surviving permutation. Critically, permutation prevented systematic improvements over the null that characterize our main results (InfoX values 0.05 to 0.20), indicating that the decoders measure spatially specific information transfer rather than generic feature statistics.

\section{Hub Marginalization Dynamics}
\label{app:hub_dynamics}

This appendix provides complete layer-wise dynamics of hub marginalization and consensus formation across all three model scales, supporting Section~\ref{sec:mechanism}. We track two metrics: \textbf{Attention Consensus Index (ACI)} $= 1 - \sigma_2$, where $\sigma_2$ is the second largest singular value of the symmetrized attention transition matrix (higher values indicate faster global mixing), and \textbf{CLS Centrality (CCC)} defined as the stationary distribution probability mass on the \texttt{[CLS]} token (lower values indicate hub marginalization).

Table~\ref{tab:hub_dynamics_full} presents complete layer-wise metrics. All models begin with high CCC (0.374 to 0.502), indicating strong initial hub centrality, then progressively marginalize to different final values while ACI generally increases. Critical transitions occur at model-specific pivot points: ViT-B shows pronounced reorganization during layers 8 to 10 (CCC drops 43\% while ACI increases 35\%), coinciding with the Information Pivot. ViT-L exhibits instability at layer 20 where CCC spikes to 0.049 from 0.004, potentially explaining its suboptimal geometric quality despite achieving the highest final ACI (0.955).

\begin{table}[!htbp]
\centering
\caption{Layer-wise hub marginalization dynamics. Values shown are averages across the validation set.}
\label{tab:hub_dynamics_full}
\small
\begin{tabular}{lccccccc}
\toprule
& \multicolumn{2}{c}{\textbf{ViT-S/16}} & \multicolumn{2}{c}{\textbf{ViT-B/16}} & \multicolumn{2}{c}{\textbf{ViT-L/16}} \\
\cmidrule(lr){2-3} \cmidrule(lr){4-5} \cmidrule(lr){6-7}
\textbf{Layer} & ACI & CCC & ACI & CCC & ACI & CCC \\
\midrule
0  & 0.634 & 0.458 & 0.738 & 0.502 & 0.889 & 0.373 \\
1  & 0.371 & 0.179 & 0.523 & 0.310 & 0.618 & 0.260 \\
2  & 0.376 & 0.211 & 0.432 & 0.310 & 0.575 & 0.297 \\
3  & 0.522 & 0.130 & 0.540 & 0.273 & 0.518 & 0.265 \\
4  & 0.474 & 0.080 & 0.520 & 0.305 & 0.726 & 0.332 \\
5  & 0.491 & 0.090 & 0.487 & 0.147 & 0.608 & 0.360 \\
6  & 0.528 & 0.047 & 0.598 & 0.099 & 0.634 & 0.254 \\
7  & 0.556 & 0.019 & 0.566 & 0.053 & 0.621 & 0.289 \\
8  & 0.662 & 0.009 & 0.671 & 0.007 & 0.571 & 0.184 \\
9  & 0.800 & 0.016 & 0.824 & 0.005 & 0.574 & 0.126 \\
10 & 0.866 & 0.015 & 0.905 & 0.004 & 0.642 & 0.074 \\
11 & 0.937 & 0.018 & 0.914 & 0.004 & 0.612 & 0.070 \\
12 & -- & -- & -- & -- & 0.595 & 0.047 \\
13 & -- & -- & -- & -- & 0.688 & 0.035 \\
14 & -- & -- & -- & -- & 0.735 & 0.024 \\
15 & -- & -- & -- & -- & 0.754 & 0.007 \\
16 & -- & -- & -- & -- & 0.799 & 0.005 \\
17 & -- & -- & -- & -- & 0.839 & 0.004 \\
18 & -- & -- & -- & -- & 0.892 & 0.002 \\
19 & -- & -- & -- & -- & 0.904 & 0.004 \\
20 & -- & -- & -- & -- & 0.939 & 0.049 \\
21 & -- & -- & -- & -- & 0.949 & 0.010 \\
22 & -- & -- & -- & -- & 0.984 & 0.001 \\
23 & -- & -- & -- & -- & 0.955 & 0.010 \\
\bottomrule
\end{tabular}
\end{table}

The hub marginalization patterns correlate strongly with Information Scrambling Index (Section~\ref{sec:information_plane}): ViT-S's collapsed scrambling (-0.001) yields incomplete marginalization (CCC=0.018); ViT-B's controlled scrambling (0.009) enables near-complete marginalization (CCC=0.004); ViT-L's excessive scrambling (0.031) coincides with unstable marginalization. These correlations suggest scrambling and marginalization are facets of the same computational reorganization.

\end{document}